%% file: main.tex
\documentclass[letterpaper, 10 pt, conference]{ieeeconf}  
\IEEEoverridecommandlockouts                              
\usepackage[dvipsnames]{xcolor}
\usepackage{graphicx}
\usepackage{cite}
\usepackage{graphics} 
\usepackage{epsfig} 
\usepackage{times} 
\usepackage{amsmath} 
\usepackage{amssymb}  
\usepackage{subcaption}
\usepackage[ruled]{algorithm2e}
\usepackage{siunitx} 
\usepackage{booktabs} 
\usepackage{makecell}
\usepackage{multirow}
\usepackage{color, colortbl}
\usepackage{adjustbox}

\definecolor{Gray}{gray}{0.9}

\graphicspath{{fig/}}
\title{\LARGE \bf
Multi-Step Recurrent Q-Learning for Robotic Velcro Peeling}
\author{Jiacheng Yuan$^{1}$, Nicolai H{\"a}ni and Volkan Isler$^{2}$
\thanks{$^{1}$ is with Department of Electrical and Computer Engineering,
        University of Minnesota, Minneapolis, MN, 55455, USA
        {\tt\small yuanx320@umn.edu}}%
\thanks{$^{2}$ are with the Department of Computer Science and Engineering, University of Minnesota, Minneapolis, MN, 55455, USA
        {\tt\small haeni001@umn.edu, isler@umn.edu}}%
}

\begin{document}
\maketitle
\thispagestyle{empty}
\pagestyle{empty}

\input{0-abstract}
\input{1-intro}
\input{2-relwork}
\input{3-background}
\input{4-formulation}
\input{5-experiments}
\input{6-results}

\input{7-conclusion}

\bibliographystyle{IEEEtran}
\bibliography{velcro.bib}

\end{document}

%% file: 0-abstract.tex
\begin{abstract}
Learning object manipulation is a critical skill for robots to interact with their environment. Even though there has been significant progress in robotic manipulation of rigid objects, interacting with non-rigid objects remains challenging for robots. In this work, we introduce velcro peeling as a representative application for robotic manipulation of non-rigid objects in complex environments. We present a method of learning force-based manipulation from noisy and incomplete sensor inputs in partially observable environments by modeling long term dependencies between measurements with a multi-step deep recurrent network. We present experiments on a real robot to show the necessity of modeling these long term dependencies and validate our approach in simulation and robot experiments. Our results show that using tactile input enables the robot to overcome geometric uncertainties present in the environment with high fidelity in $\sim 90\%$ of all cases, outperforming the baselines by a large margin.
\end{abstract}

%% file: 1-intro.tex
\section{Introduction}
\label{sec:intro}
\begin{figure}[htbp!]
    \centering
    \includegraphics[width=0.8\linewidth]{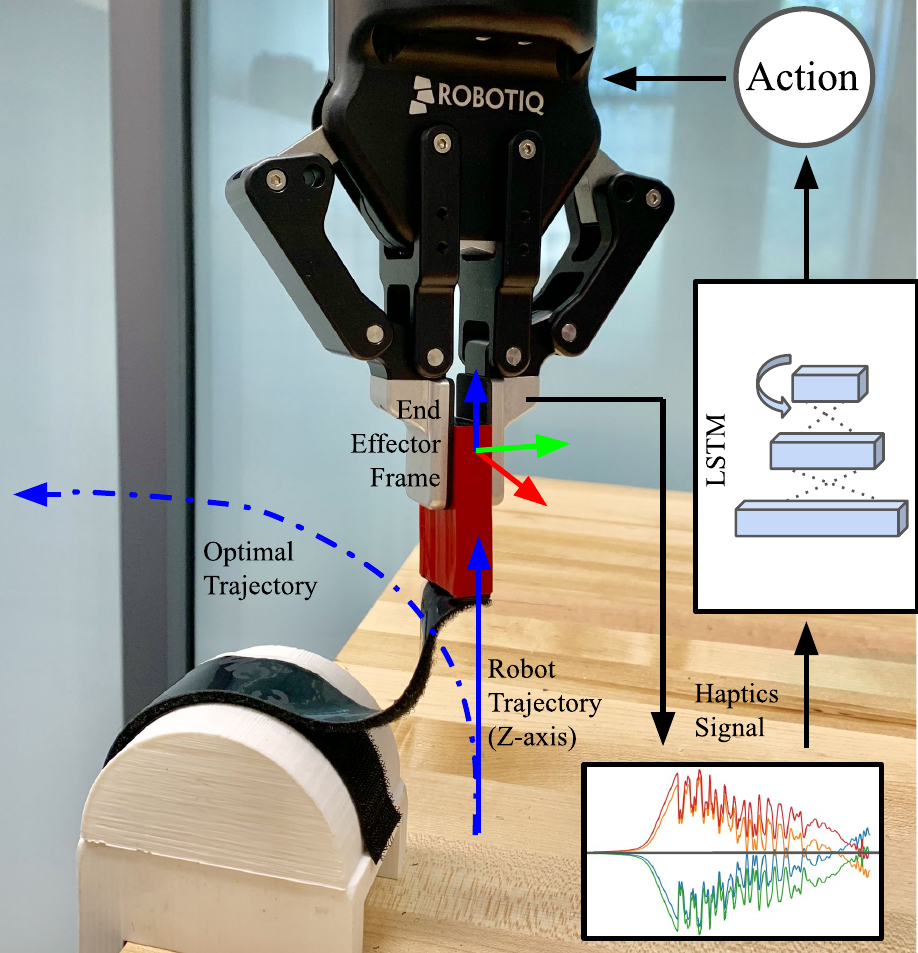}
    \caption{\small{We introduce velcro-peeling over complex surfaces as a representative task for manipulating non-rigid objects using tactile feedback.}}
    \label{fig:concept}
\end{figure}

Manipulation enables robots to physically interact with their environment. Robotics researchers have made significant progress on tasks such as grasping~\cite{shimoga1996robot,pinto2016supersizing,grasp_unseen,levine2018learning,zeng2018learning} and dexterous manipulation~\cite{yousef2011tactile,andrychowicz2020learning} of {rigid} objects. In this work, we focus on the problem of interacting with \emph{non-rigid} objects. Learning to manipulate non-rigid objects allows robots to handle fragile~\cite{wen2020force,soft_object} and flexible objects~\cite{rambow2012autonomous}, or household items~\cite{matas2018sim,wu2019learning}. Although research on rigid object manipulation is a mature field, existing techniques can not be applied directly to non-rigid objects~\cite{sanchez2018robotic}.  
In this paper, we introduce velcro peeling as an illustrative application for manipulating a non-rigid object in a complex geometric setting (Fig.~\ref{fig:concept}).

The goal is to peel velcro over a surface with unknown geometry, which provides a unique set of challenges: (1)~force feedback measurements can be ambiguous, (2)~visual feedback is not always viable due to self-occlusion, and (3)~the system state is not directly observable. As a camera's view can be blocked through self-occlusion and merging different sensor modalities, such as vision and touch, is challenging, we investigate if a robot can learn to peel velcro from force feedback alone. However, relying solely on touch brings with it additional challenges. Suppose the robot established an initial grasp of the velcro endpoint, and the task is to open the velcro fully. Tactile feedback can only be measured if the material resists the robot's pulling motion. However, we show in Section~\ref{sec:background} that the measured feedback signal of peeling a velcro is nearly identical to when the robot is pulling in the wrong direction.
Additionally, our experiments on a real robot identified a force void space, where no force feedback can be measured at all. Learning correct behavior from such sensor signals requires reasoning capabilities over a long time horizon. 
We believe that velcro-peeling provides an approachable, representative application for manipulating non-rigid objects with only touch feedback. It also has practical applications considering that velcro is a common material found on everyday objects such as coats, bags, and shoes.

We propose a novel simulation environment where a velcro strip is placed on a variety of surfaces, including planar and cylindrical ones (Figure~\ref{fig:concept}). We show how the ideal, fully-observable version of the peeling task can be formulated as a Markov Decision Process and solved optimally. If only tactile measurements are available, the problem becomes partially observable. For this version, we present a multi-step Deep Recurrent Network (DQRN) that can successfully solve more than 90\% of the configurations under geometric uncertainty and ambiguous sensor feedback. Our method improves performance over existing baselines by over 20\%.

Our contributions are summarized as follows: 
\begin{itemize}
   \item We introduce velcro peeling as a representative application to learn non-rigid robotic object manipulation from only touch feedback.
    \item We present a Multi-Step DRQN network that handles long-term dependencies between sensor measurements to peel velcro strips from varying geometric shapes successfully.
    \item We validate our approach in simulation and in experiments on a real robot.
\end{itemize}

%% file: 2-relwork.tex
\section{Related Work}
\label{sec:rel_work}
Markov Decision Processes (MDP) provide a mathematical formulation for reinforcement learning problems. 
 The standard MDP formulation assumes that the current state of the environment is fully observable and that the optimal action choice depends solely on this current state. However, estimating the current state is non-trivial. As such, various neural network architectures have emerged as powerful tools to learn state estimation from observations.
For Atari gamers, Deep Q-Networks (DQN) achieved human-level control for discrete~\cite{mnih2015dqn} and continuous action spaces~\cite{dqn_continuous}. The same approach can be used for robotic manipulation. However, the MDP modeling approach performs well only in fully observable systems, such as Atari games, in which low dimensional features can be extracted from observations due to relatively simple physical environment settings. Extending these methods to manipulation tasks in 3D Cartesian space with combinations of multi-modal sensory inputs such as vision, tactile, and proprioceptive data~\cite{making_sense_of_VaT} is an active area. If the environment additionally contains uncertainty or noise, the MDP based modeling approach fails. 
Instead, the problem can be reformulated as a Partially Observable Markov Decision Process (POMDP). Two recent papers~\cite{grasping_pomdp_2007, pomdp_lite_grasping} used this approach to learn robust robotic grasping. Glashan et al.~\cite{grasping_pomdp_2007} simplified the grasping observation and state-space to discrete abstractions to reduce the probability model complexity in their POMDP formulation. Similarly, Chen et al.~\cite{pomdp_lite_grasping} fixed the hidden state variables and introduced more deterministic properties to the Bayesian transition model. They successfully achieved robust two-finger grasping under uncertainty. In more challenging manipulation tasks, where multiple objects interact with each other through collision and friction, it is non-trivial to design an optimal state-space representation.  Sung et al.~\cite{diode_deepRNN} proposed a variational Bayesian model to learn the continuous state representation from the tactile signal followed by a planning network. In contrast, we propose a method to skip extracting features altogether and learn a mapping from a tactile signal sequence to the optimal action. 

Several prior works have applied hand-designed controllers in combination with tactile feedback to solve rigid body manipulation tasks~\cite{haptic_feedback_control_grasp,inverse_sensor_modeling}.  When the geometric parameters of the object are known, a PID controller can even address the peg-in-hole problem robustly~\cite{geometric_peg-in-hole}. These controllers address particular tasks well when the states in their models are accurately measured or estimated. Koval et al.~\cite{koval2017manifold} demonstrated particle filtering for the states of a noisy robot arm informed by the tactile sensor.
Platt et al.~\cite{platt2011using} applied Bayesian estimation using tactile feedback to localize flexible materials during manipulation. Sutanto et al.~\cite{latent_space_tactile_servoing} developed an approach to predict actions to perform tactile servoing based on a learned latent space representation. The same method can also be applied to estimate object physical properties such as elasticity and stiffness~\cite{latent_elasticity,latent_stiffness}.  However, there might not exist a correlation between latent representation changes and agent actions. That is one of the cases where tactile input for feature extraction performs worse compared to other sensory inputs, for example, vision.  In the velcro peeling case, it is hard to estimate the velcro's loop and hook status based on the pressure mapping and shear force readings at the gripper finger.
Additionally, the detachment of the hook and loop introduces noise to the sensor. The resistance force could result in ambiguous sensor readings. We show that we can overcome these challenges with our proposed Multi-step DRQN architecture that considers the long term dependencies between individual observations.

Tactile sensory input can provide useful perceptional capabilities to assist manipulation~\cite{making_sense_of_VaT}. When combined with visual input, a neural network can combine the two signals and extract features for object classification. Combining vision and tactile observations were shown to benefit some applications~\cite{cross-modal_touch_and_vision,haptic_SVM_manipulation, haptics_drilling_carving}, such as slicing, drilling, and carving, which require direct perception around the contact area. In ~\cite{whats_in_container}, the authors have shown that vision and tactile feedback can identify objects inside a container individually. In our work, vision feedback does not provide a stable input signal due to self-occlusions. Instead, we show that a high-fidelity model can be learned from tactile observations only.

%% file: 3-background.tex
\section{Velcro Peeling} 
\label{sec:background}
\begin{figure}
    \centering
    \includegraphics[width=\linewidth]{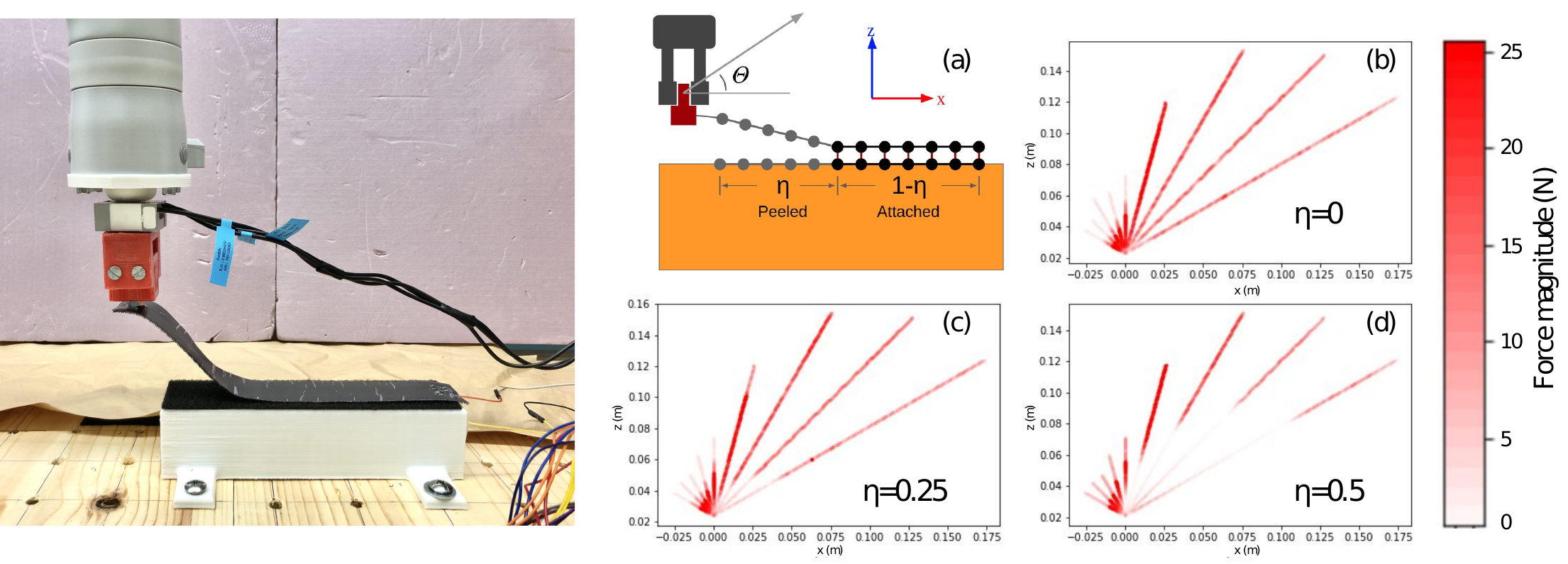}
    \caption{\small{Tactile force magnitude along fixed trajectories with varying angle $\theta$ and with different velcro initial conditions $\eta$ (percentage of the velcro that's already peeled). The robot manipulator encounters a tactile void space as more of the velcro strip is peeled. The figure is best viewed in color.}}
    \label{fig:theta_peel}
\end{figure}

This section presents an initial experiment on a real robot that highlights the challenges when designing velcro peeling strategies for complex geometries. 
We measure the force feedback while moving the manipulator along predefined trajectories, parametrized by $\theta$, the angle between the peeling direction and the x-axis. Fig.~\ref{fig:theta_peel} shows the force magnitude of each trajectory when we vary the percentage of already peeled velcro ($\eta \in [0,1])$. The force is measured with a tri-axis load sensor on a Kinova Gen 3 7-DOF manipulator's wrist, and the magnitude is indicated by the strength of the marker in the figure. The darker the dot, the larger the force magnitude. 

We observe a tactile void space (where force feedback is weak) that increases as more of the velcro is peeled off.  These extended sequences without useful feedback pose a significant challenge for standard Deep Q-learning approaches~\cite{mnih2015dqn}.  As the force feedback becomes weaker over time, the learned policy can not differentiate between states, and all Q-values become equally likely, meaning that no optimal action exists. 
In the next section, we present our approach to address this challenge.

%% file: 4-formulation.tex
\section{Method}
\label{sec:formulation}
In this section, we briefly describe our model of the velcro peeling task in simulation together with the mathematical notation (Table~\ref{tab:table1}). Next, we show that if the full state is observable,  velcro peeling can be solved by choosing greedy decisions based on the peeling boundary region. Finally, we study the partially observable cases, where the controller observes only vision or tactile feedback.

\subsection{Problem Formulation}
We aim to peel a velcro strip of uniform width, applied on surfaces of varying geometry. The manipulator initially holds the handle in hand. We discretize the velcro into $n$ consecutive flat pieces, each containing a binary value on whether the segment is peeled (1) or attached (0). The global state of the velcro represents a string of $n$ attached/peeled bits. Given tactile and proprioceptive observations from the agent, our goal is to find an optimal strategy to control the gripper to successfully change the velcro attachment state from $\{0,\ldots, 0\}$ to $\{1,\ldots, 1\}$. 
\begin{table}[ht]
    \centering
    \caption{\small{Summary of Notations}}
    \label{tab:table1}
    \begin{tabular}{l l} 
        \toprule 
        Notation & Description  \\
        \midrule
        $(d_{x}, d_{y})$   & \makecell*[{{p{6.5cm}}}]{Velcro origin translation in x,y direction in the world coordinate frame, sampled uniformly $\mathcal{U}(-1, 1)$.} \\
        $(\theta_{x}, \theta_{y}, \theta_{z})$ & \makecell*[{{p{6.5cm}}}]{Velcro rotation with respect to x,y,z axis of the world coordinate frame, $\mathcal{U}(-\pi, \pi)$ }  \\ 
        $r$          & \makecell[{{p{6.5cm}}}]{Table radius for cylindrical velcro, $\mathcal{U}(0.4, 0.8)$ } \\
        \midrule
        ${O_{t}}$ & \makecell*[{{p{6.5cm}}}]{Force, torque and position observation of the gripper at time step $t$}\\
        ${S_{t}}$ & \makecell*[{{p{6.5cm}}}]{State of the discrete velcro model at time step $t$, including all nodes' and spring damper units' position and velocity} \\
        ${A_{t}}$ & \makecell[{{p{6.5cm}}}]{Discrete action set of the agent. Moves the gripper for a fixed distance up, down, forward, backward, left or right}                      \\
        $T$     &   \makecell*[{{p{6.5cm}}}]{State transition function $T: S_{t} \to S_{t+1} | A_{t}$ } \\
        $R$     & \makecell[{{p{6.5cm}}}]{Reward function containing the number of bits flipped from $0$ to $1$ in the velcro attachment state}\\
        \bottomrule
    \end{tabular}
\end{table}

\subsection{Simulation Model}
The velcro strip is simulated as a 2-D net of point mass nodes connected by spring-damper units. Variables relevant to the state (listed in Table~\ref{tab:table1}) are $d, \theta, r$ representing translation, rotation, and radius of the surface shape on which the velcro is mounted. The velcro strips hooks and loops are simulated with breakable tendons. At time step $t$, the environment state $S$ includes the position and velocity of each velcro node, the length of all tendons, the manipulator's end-effector pose, and the tactile feedback measurement. Additional details of the simulation setup are introduced in section~\ref{sec:experiments}.

\subsection{Our Approach}
Humans peel a velcro strip by grasping one end and pulling towards a direction roughly between the surface tangential and the peeling boundary's surface normal. In our experiments in Section~\ref{sec:experiments}, we show that if the states of all velcro nodes are observable, we can compute the surface tangential and normal. In this case, a simple greedy strategy suffices to peel the velcro.

Of course, in real-world environments, it is not possible to observe these environment variables directly. Additionally, it is challenging to estimate the velcro's geometric properties accurately. Since the environment state is not fully observable, we formulate our approach as a  Partially Observable Markov Decision Process (POMDP). A POMDP is characterized by a tuple of 6 values: States $\mathcal{S}$, actions $\mathcal{A}$, a state transition function $\mathcal{T}$, reward function $\mathcal{R}$, and observations $O$ according to an observation function $\Omega$. In our case, the observations $O \in \Omega$ contain only the position of the end-effector and the tactile feedback measurement. We use the area of peeled velcro as our reward $R$, and we define six possible manipulator actions (move left, right, forward, backward, up and down).

We use reinforcement learning~\cite{sutton2018reinforcement} to learn a control policy $\pi$ that at each time-step $t$ receives observation $O_t$ chooses an action $a_t \in \mathcal{A}$ that optimizes our long term reward. If the state $s_t$ is directly observable, the problem can be solved by learning an optimal policy $\pi*$ that maximizes the expected sum of future rewards, given by $R_t = \sum_{i=t}^\infty \gamma R(s_i, s_{i+1})$, i.e. $\gamma$-discounted sum over an infinite time horizon of future returns. Q-Learning~\cite{watkins1992q} is a model-free off-policy algorithm to estimate these expected long term rewards or Q-values. However, in real-world scenarios, it is often impossible to observe the state $s_t$ directly~\cite{hausknecht_deep_2015}. In this case, estimating the Q-values from a single observation can be arbitrarily bad, since $Q(s_t, a_t | \theta) \neq Q(o_t, a_t|\theta)$. Hausknecht and Stone~\cite{hausknecht_deep_2015} showed that estimating the state using multiple observations together with a Deep Recurrent Q-Network (DRQN) leads to better policies in partially observed environments. 

\begin{figure}[ht!]
    \centering
    \includegraphics[width=\linewidth]{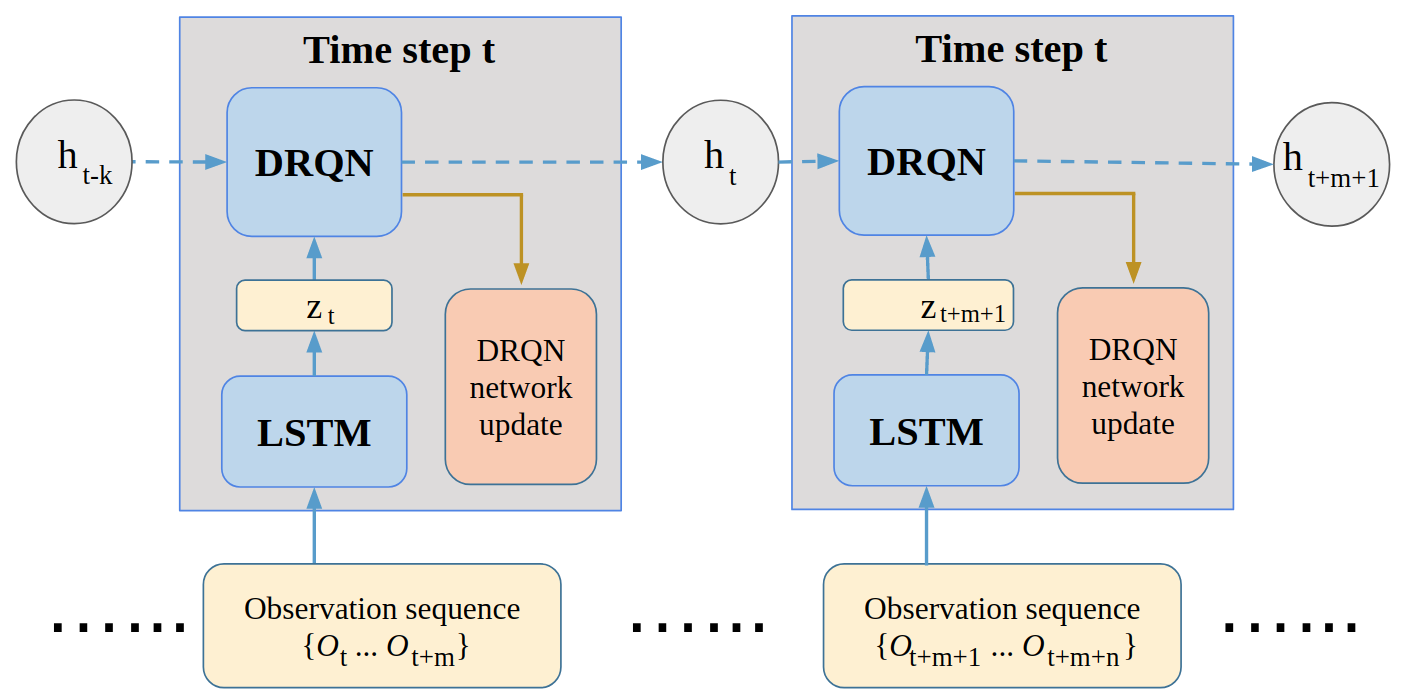}
    \caption{\small{Proposed architecture of our Multi-step DRQN network. We use an LSTM layer to learn a fixed size tactile feedback representation as input to the DRQN.}}
    \label{fig:net_struct}
\end{figure}

However, in our experiments in Section~\ref{sec:experiments}, we show empirically that the standard DRQN approach suffers from the long time-scale memory transport problem~\cite{longtime_mem_transpt}. We show that we can address this issue by slowing down the Q-value estimation frequency. To achieve this, we propose a Multi-Step DRQN approach, as shown in Fig. \ref{fig:net_struct}.
When the tactile feedback is weak (usually when there is slack in the peeled part of the velcro), the force feedback provided contains little meaningful information. In our early experiment in Section~\ref{sec:background}, we showed that this tactile void space grows as the percentage of peeled velcro increases. To estimate the Q-values within this void space, the agent needs to reason from observations that span a long time horizon. Our multi-step DQRN outputs a single action for a maximum of $k$ time steps if tactile feedback is weak to overcome this issue. If sensor measurements are reasonable, the agent predicts the action-state value and chooses the next action in the standard DRQN network. Since the resulting observations do not have a fixed length, we propose using a Long-Short Term Memory (LSTM) layer along with two linear layers to learn a fixed-sized tactile feature vector. This tactile feature vector is then used as input for the DRQN network to estimate the Q-values. Our proposed network architecture enforces more state explorations inside the tactile void space and slows down the Q-value estimation frequency. We also applied the Double Q-learning method from~\cite{double_dqn}, which uses two independent Q-networks for Q-value estimation to increase training stability.

%% file: 5-experiments.tex
\section{Experiments: Design and Setup}
\label{sec:experiments}
In this section, we first introduce the velcro strip simulation model and associated parameters. Next, we introduce the evaluation metrics, implementation details, and the evaluation baselines. Finally, we introduce our setup for real-world evaluation.
\begin{figure}[ht!]
    \centering
    \includegraphics[width=0.8\linewidth]{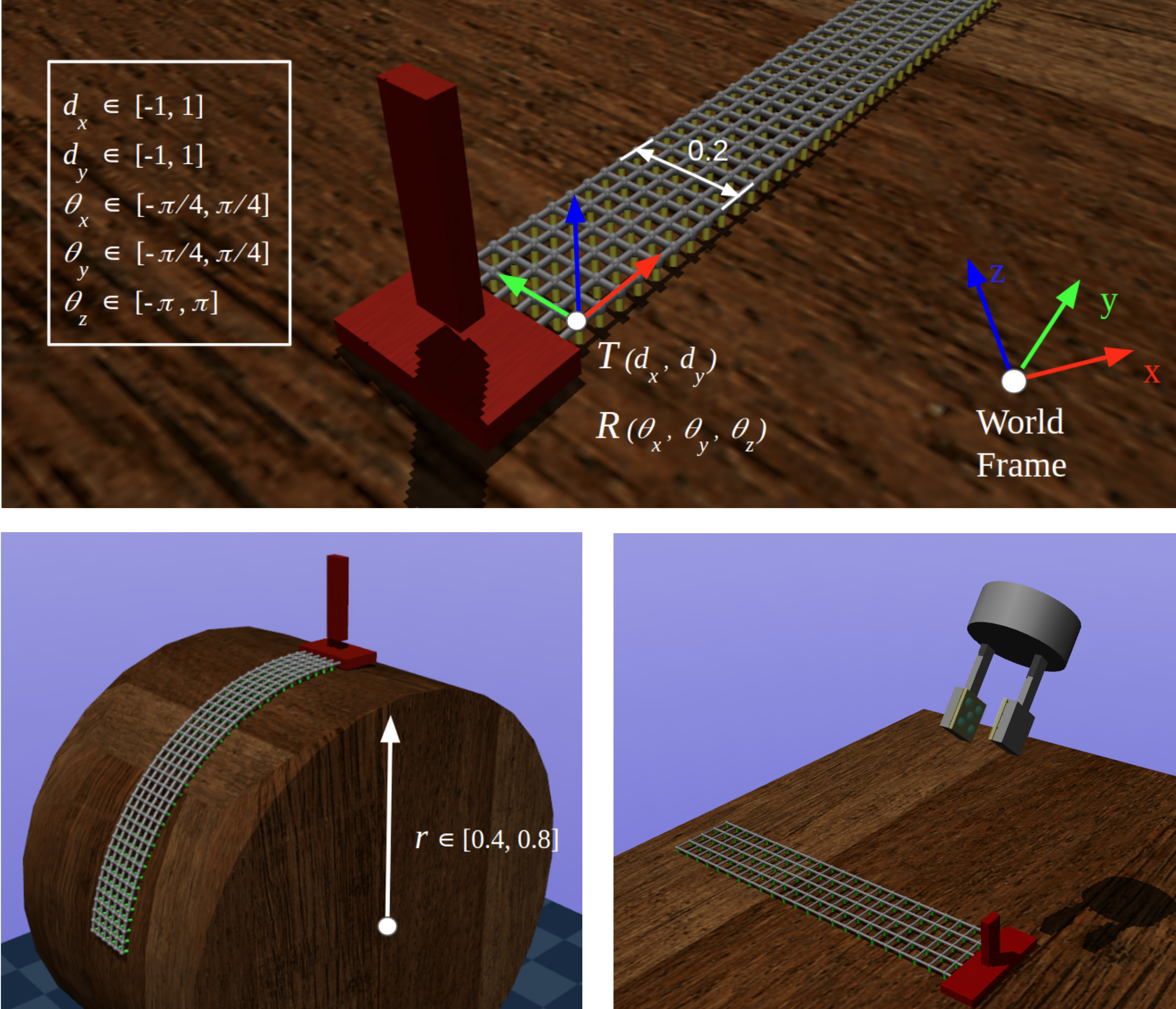}
    \caption{\small{\textbf{top \& bottom left: }Velcro geometric uncertainty parameters. \textbf{bottom right: }Floating gripper in simulation}}
    \label{fig:sim}
\end{figure}
\subsection{Velcro Model} 
We use MuJoCo, a fast and accurate physics engine optimized for dynamical systems with rich contacts and constraints for simulations.
We simulate the velcro strip as closed-loop kinematic chains, consisting of two 2D arrays of point mass nodes connected through tendons. Each tendon is modeled as a spring-damper unit to impose force constraints and motion limits (Fig. \ref{fig:sim}). The bottom layer is fixed on a table structure, while the upper layer is constrained only by the tendons. Once the spring tension exceeds a certain threshold, we reduce the spring constant to 0 to eliminate the relevant constraint, i.e., the hook detaches from the loop. 
To simplify simulation behavior, we do not recover the spring constant even if the spring displacement goes back to zero, i.e., the hook does not reattach to the loop once it was detached. The tendon spring constant remains unchanged throughout the training for consistent dynamic behavior.

\subsection{Model Generation}
To introduce geometric variations, we parameterize three different geometry scenarios from which we sample training data. 
Namely, we sample models with variations of (Fig. \ref{fig:sim}): velcro translation ($d_{x}$, $d_{y}$), rotation ($\theta_{x}$, $\theta_{y}$, $\theta_{z}$) and concavity, where the velcro is generated on a cylindrical surface with radius $r$. During training, we randomly initialize a model by choosing a set of parameters for each episode.

\subsection{Robot Agent}
We simulate only the gripper part of the manipulator to achieve fast planning. The gripper consists of a standard parallel jaw gripper equipped with force-torque sensors at the fingertips (Fig. \ref{fig:sim}). We remove the computation complexity from inverse kinematics by controlling the gripper directly in the end effector frame using position and velocity commands.

Unlike~\cite{making_sense_of_VaT}, we use a set of discrete actions in 3D Cartesian space. Each Cartesian space action displaces the manipulator by a fixed $\delta$. The action displacement is sampled into a quintic polynomial trajectory to get the joint position and joint velocity command at each time step through a PD controller. The Cartesian space action magnitude is fixed to ensure an equal number of simulation steps.

\subsection{Reward design}
In our modeling approach, the velcro tendon spring constant is set to 0 to approximate the hooks' detaching from the loop. We call this process breaking the tendons. The reward is assigned by how many tendons the manipulator breaks during a single step.

\subsection{Evaluation Metric}
\label{subsec:evaluation}
We evaluate our approach in environments generated similarly to the models used during training. To demonstrate short-comings and failure-cases occurring during geometric self-occlusion, we generate three test cases with different parameters. Test case 1 includes variation in translation and rotation $\theta_{z}$. Test case 2 additionally contains variations in the other two rotation parameter $\theta_{x}$ and $\theta_{y}$. Test case 3 uniformly samples 50\% of the environments on planar- and 50\% on to cylinder-shaped table whose radius is controlled by the parameter $r$. In total, we generate 500 examples for test cases 1 and 2, and 1000 samples for case 3. We measure success via completion ratio (number of broken tendons compared to the total number of tendons). To discourage infinite exploration, we formulate three termination criteria: \textit{success} in cases where the manipulator peels off the whole velcro, \textit{failure} in cases where the manipulator loses hold of the velcro, and \textit{failure} in cases where the time limit is exceeded. We set the time limit to 200 steps for both training and testing. 

\subsection{Implementation Details}
Our tactile sequence network is implemented as an LSTM layer, followed by two linear layers. In the simulation, the tactile input consists of 3 values for force and torque for both fingers and runs at $30Hz$. Thus each observation contains 186 values (6 values for end-effector pose and 180 for tactile observation). 
The output hidden vector $z$ is of size 150. 
Our Q-network is a three-layer Multilayer Perceptron (MLP), with two linear layers followed by an LSTM layer. Finally, a linear layer outputs a Q-value for each action. We use ReLU as non-linearities and batch-normalization layers for weight normalization. During training, we jointly learn the linear layer weights and LSTM weights. 
For each episode, we randomly sample the geometric uncertainty parameters to generate a new model, and the episode terminates after 200 action steps. The policy is trained for 1500 episodes using RMSProp optimizer with a learning rate of $2e^{-5}$ for about 30 hours on a single NVIDIA GeForce GTX 1080.

\input{baselines}

\subsection{Real Robot Setup}
For real-world evaluations, we use the same setup as described in \ref{sec:background}. We fix the 3D printed structure on the table and rotate the robot's end-effector frame at initialization, rotating the discrete cartesian action space and the tactile force frame accordingly. We then add a random offset to the position observation to introduce uncertainties for translation.
We found that the hook and loop detaching during the velcro peeling process yields a unique audio signal that can approximate the reward signal. We sample an audio signal of a single, noncontinuous velcro peeling process in a quiet room for $30s$. We then threshold and average the spectrum of this signal over time to filter in the frequency domain. A 1D convolution with the signal spectrum yields a filtered signal without background noise, which we use to approximate the amount of velcro peeled during the process.

We train the real robot agent for 200 episodes and then test the result on test cases 1 and 2. In total, we generated 30 examples for both case 1 and case 2.


%% file: baselines.tex
\subsection{Baseline Strategies}
\begin{figure}[htbp!]
    \centering
    \includegraphics[width=\linewidth]{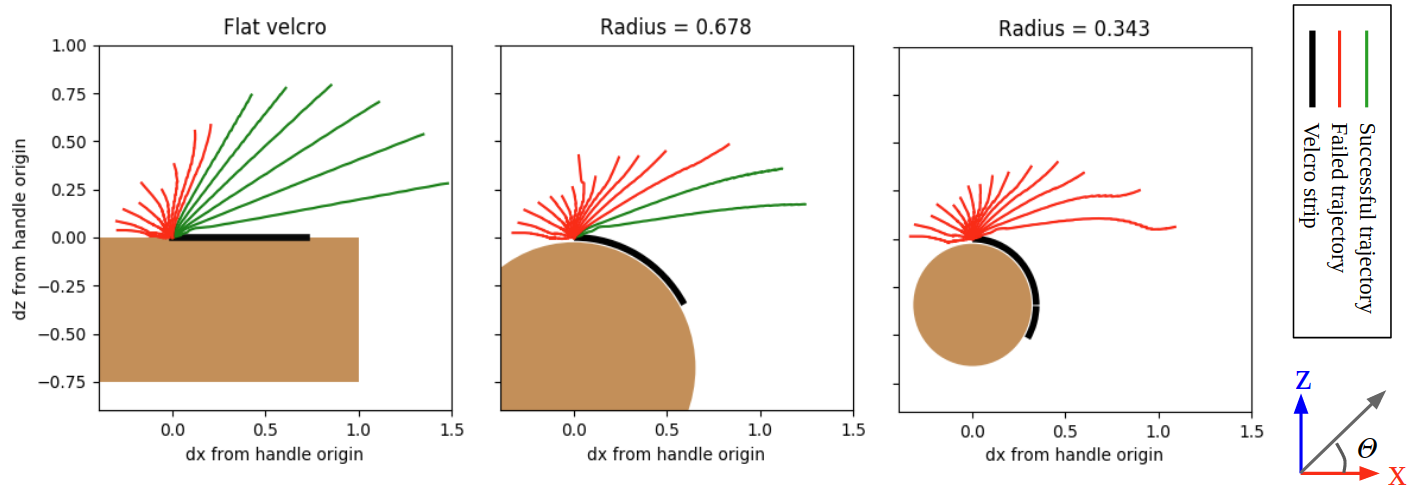}
    \caption{\small{Cross section view of a series of end effector trajectories with different table geometries. Table (brown) has varying radii, and the velcro strips (thick black curves) are applied on the table. The trajectories (red for fail and green for success) are parametrized by angle theta (pulling direction).}}
    \label{fig:theta_radius}
\end{figure}

We compare our approach with six different baselines.
\textbf{Full Observation}:
\textbf{Geometric input greedy approach (Geom-greedy)}
If the full state $S$ is observable, a hand-designed strategy based on the peeling boundary's geometry information is sufficient to solve the peeling task. Specifically, the peeling orientation, the position and normal vector at the peeling boundary, and the gripper orientation need to be observed. We use this information as the basis for a greedy algorithm similar to the one presented in Section~\ref{sec:background}. The end effector follows trajectories defined by $\theta$ (see Fig.~\ref{fig:theta_radius}). The trajectories that yield successful peels are plotted in green while failed ones are shown in red. The results show that by approximating the velcro shape with straight line segments, the agent only needs to drive the end effector inside a peeling cone while increasing the cone origin's distance.        
\textbf{Partial Observation}
As geometric velcro properties can not be observed directly in the real world, we limit the observations to measurable data: vision or tactile input.
\textbf{Open loop (sweep-through)}
The most straightforward strategy is to choose to pull in a single direction. We randomly sample a direction from a hemisphere, assuming the velcro strip's topside is always facing up ($+z$ direction). The gripper moves towards the sampled direction until failure or success. 
    
\textbf{State Estimation + Hand-designed Strategy}
The state estimation + Hand-designed strategy estimates the peeling boundary and normal of the velcro using either vision or tactile input and generates actions using the previously presented Geom-greedy approach. We design two strategies based on the sensor modality: Vision-greedy and Tactile-greedy.
For the Vision-greedy method, we place a camera that captures RGB-D images in the simulation environment. We use a neural network with a ResNet-18 encoder followed by three additional linear layers to predict the peeling orientation, the position, and the normal vector at the peeling boundary from images. We randomly explore the state space and collect 4532 images and the associated geometric features from 1000 different velcro configurations. 
For the Tactile-greedy method, we sampled 1000 random exploration trajectories and trained a recurrent neural network to predict the peeling boundary's geometry information.
The Vision-greedy and Tactile-greedy methods are trained in a supervised manner until convergence.  
        
\textbf{Reactive policy}
The reactive policy network is a standard Q-value network. This network contains no memory of previously chosen actions/observations and selects the next action based on the current observation.
    
\textbf{Single-step DRQN}
The single-step DRQN closely follows the approach introduced by~\cite{hausknecht_deep_2015}. The current observation is processed by a DRQN network to predict Q-values at every step. The network keeps memory in the DRQN's internal hidden state.

%% file: 6-results.tex
\section{Results}
\label{sec:results}
We record the episodic return (total number of tendons broken), time (number of discrete action steps), and final result as success or failure. The episodic returns are discretized into 5 ranges: $0\%-20\%$, $20\%-40\%$, $40\%-60\%$, $60\%-80\%$ and $80\%-100\%$. Fig. \ref{fig:bar_charts} shows the 100\% stacked column charts on the episodic returns. 
\input{result_bar}
\input{result_table}

\input{result_ablation}

We also compute the success rate, average time step, and average episodic return for each test case and summarize them in Table~\ref{tab:resulttable1}. In full observability, the Geom-greedy approach achieves a 100\% success rate with the shortest peeling time among all methods. The Geom-greedy's success in the fully observable case is expected, as the Geom-greedy approach can follow the direction between surface tangential and the normal of the peeling boundary throughout the episode to achieve success. When we compare the Vision-greedy approach with the Geom-greedy, we see that the Vision-greedy approach achieves perfect performance test 1.  However, when the test cases contain geometric self occlusions, as in test cases 2 and 3, the performance decays quickly. Among all approaches with partial observation, our Multi-step DRQN achieved the highest success rate on test sets 2 and 3 and competitive performance on test set 1. 

To show the importance of tactile feedback measurements, we provide ablations of the used observation representation. The agent can observe only the gripper position and the tactile feedback, and the training is conducted with both Single-step and Multi-step DRQN formulations. Additionally, we study the influence of the sample time steps $\tau$ on the DRQN agents. Both ablation study results are summarized in Table~\ref{tab:resulttable1} and Table~\ref{tab:ablation2}. 

In Table~\ref{tab:resulttable1}, both single-step and multi-step DRQN approaches show performance drop when either force or position input is missing, indicating that observing only part of the geometry is not sufficient to solve the task, and the partly observed tactile feedback can be ambiguous regarding the underlying physical properties.
In Table~\ref{tab:ablation2}, the multi-step DRQN agent is trained with the same hyper-parameters except for $\tau$, which is selected from $\tau = {1,2,4,8}$. For the case where $\tau =1$ is identical to the reactive method. The small performance difference can be attributed to differently initialized parameters. The performance shows significant improvement when $\tau$ increases, indicating the importance of memory and long time reasoning capability for the agent to solve this task. 

In the real robot evaluations, we show that the multi-step approach can achieve a similar success rate as in simulations. It also outperforms the single-step DRQN approach both in terms of success rate and the average timestep it takes to finish the task.
\input{result_real_robot}

%% file: result_bar.tex
\begin{figure}[bp!]
	\centering
		\centering
		\includegraphics[width=\linewidth]{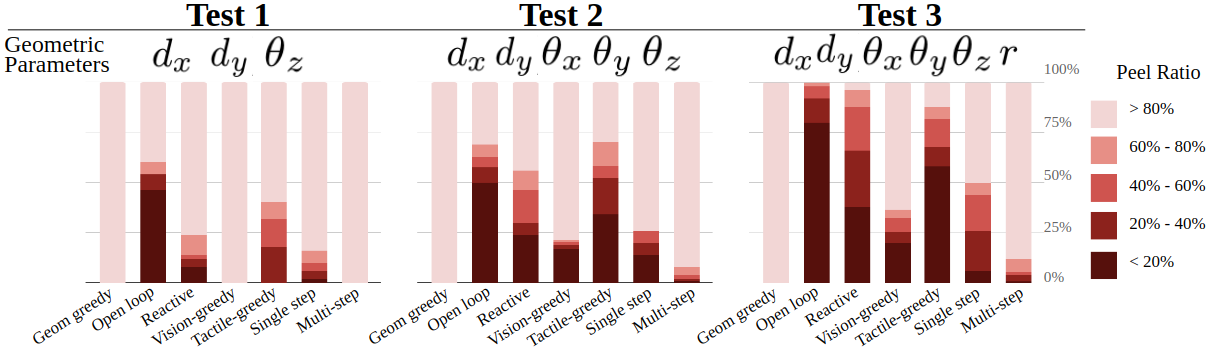}

	\caption{\small{Stacked bar charts of test on 3 test cases. Bar segments of 5 colors correspond to percent of test cases in which the agent open 0\%-20\%, 20\%-40\%, 40\%-60\% 60\%-80\% and 80\%-100\% of the velcro strip.}}
	\label{fig:bar_charts}
\end{figure}

%% file: result_table.tex
\begin{table}[htbp!]
	\centering
	\caption{\small{Performance of our approach and baselines in three test cases ($\eta$: success rate $t_{f}$: finish time $G$: episodic return ) }}
	\label{tab:resulttable1}
	\begin{adjustbox}{width=\linewidth}
		\begin{tabular}{l c c c c c c c c c} 
			\toprule 
			{} & \multicolumn{3}{c}{\bf{Test 1}} & \multicolumn{3}{c}{\bf{Test 2}} & \multicolumn{3}{c}{\textbf{Test 3}} \\
			{} & $\eta$ & $t_{f}$ & $G$ &  $\eta$ & $t_{f}$ & $G$ & $\eta$ & $t_{f}$ & $G$ \\
			\midrule
		    	\multicolumn{10}{l}{\textbf{Performance}} \\
			\midrule
    			\multicolumn{10}{l}{Full Observation}  \\
    			    \makecell[l]{\quad Geom-greedy}        & 100\%  & 49.5  & 216 & 100\%   & 52.7 & 216 & 100\% & 61.8  & 216  \\
    
    		 	\multicolumn{10}{l}{Partial Observation} \\
        		 	\makecell[l]{\quad Open loop  }        & 36\%          & 144  & 79  & 21\%   &  160  & 51  & 10\%  & 173 & 29   \\
        		    \makecell[l]{\quad Vision-greedy}      & \textbf{100\%}          & \textbf{62}   & \textbf{216}  & 77\%  & 115   & 174   & 59\%  & 164   & 152   \\
        		    \makecell[l]{\quad Tactile-greedy}     & 59\%           & 146  & 139  & 22\%  & 174   & 84    & 15\%  & 186   & 54   \\
        		    \makecell[l]{\quad Reactive}           & 62\%            & 157  & 146  & 44\%  & 171   & 118   & 22\%  & 183  & 73   \\  \makecell[l]{\quad Single-step DRQN wo pos.} & 56\%           & 158  &  127   & 39\%          & 173  & 91    & 26\%            & 165   & 75  \\
        		    \makecell[l]{\quad Single-step DRQN wo force}    & 60\%           & 171  &  145   & 57\%          & 154  & 147   & 41\%            & 170  & 103 \\
        			\makecell[l]{\quad Single-step DRQN}   & 82\%            & 139  & 184  & 68\%  & 143   & 155   & 47\%  & 152  & 122  \\
        			\makecell[l]{\quad Multi-step DRQN wo pos.} & 68\%          & 142 & 167    & 65\%          & 149  & 153   & 56\%          & 177  & 148 \\
				    \makecell[l]{\quad Multi-step DRQN wo force}    & 75\%          & 146 & 174    & 70\%          & 151  & 172   & 52\%          & 182  & 136  \\
        			\makecell[l]{\quad Multi-step DRQN}    & 98\%   & 82  & 213  & \textbf{92\%} & \textbf{97}  & \textbf{206}  & \textbf{85\%}  & \textbf{129} & \textbf{191}
		\end{tabular}
	\end{adjustbox}
\end{table}

%% file: result_ablation.tex
\begin{table}[ht]
	\centering
		\caption{\small{Ablation study of frame time step $\tau$ }}
		\label{tab:ablation2}
		\begin{tabular}{l c c c c c c c c c} 
			\toprule 
			{} & \multicolumn{3}{c}{\bf{Test 1}} & \multicolumn{3}{c}{\bf{Test 2}} & \multicolumn{3}{c}{\textbf{Test 3}} \\
			 $\tau$ & $\eta$ & $t_{f}$ & $G$ &  $\eta$ & $t_{f}$ & $G$ & $\eta$ & $t_{f}$ & $G$ \\
			\midrule
								   
			1   & 82\%            & 138  & 184    & 68\%            & 143    & 155    & 47\%            & 152    & 122 \\
			2   & 84\%            & 117  & 144    & 65\%            & 157    & 142    & 61\%            & 164    & 151  \\
			4   & 97\%            & 87   & 210    & 88\%            & 115    & 207    & 76\%            & 136    & 173  \\
			8   & \textbf{98\%}   & 82   & 213    & \textbf{92\%}   & 97     & 206    & \textbf{85\%}   & 129    & 191    \\
			\bottomrule
		\end{tabular}
\end{table}

%% file: result_real_robot.tex
\begin{table}[ht]
    \centering
    \caption{\small{Real Robot Evaluation Results}}
    \label{tab:table_real_robot}
    \begin{tabular}{c c c c c} 
		\toprule
		{} & \multicolumn{2}{c}{\bf{Test 1}} & \multicolumn{2}{c}{\bf{Test 2}} \\
		{} &\makecell{Success\\Rate} & \makecell{Average\\Timestep} &  \makecell{Success\\Rate} & \makecell{Average\\Timestep} \\
		
		\midrule
		 Open loop     & 40\%           & 18    & 27\%           & 25   \\
         Single-step   & 90\%           & 28    & 83\%           & 32  \\
         Multi-step    & \textbf{97\%}  & 21    & \textbf{88\%}  & 25    \\
		\bottomrule
	\end{tabular}
\end{table}

%% file: 7-conclusion.tex
\section{Conclusion}
\label{sec:conclusion}
We introduced the task of peeling velcro strips mounted on varying geometries as a new task for non-rigid robotic manipulation. To solve this task in the presence of environmental uncertainties, we proposed a novel Multi-step DRQN architecture that outperforms all baselines in two out of three test cases and achieves competitive performance on the last one. We provide experiments in both simulation and real robot setup to evaluate our approach and explain the need for a network that models the long term dependencies between observations. The empirical evaluation and comparison with multiple baseline methods provide a benchmark for future work to study this problem.
Exciting future research directions include implementing the velcro strip's initial grasping and generalizing to more complex geometric configurations.

\section*{ACKNOWLEDGMENT}
This work has been supported in part by the Research Council of Norway project 303607.